\documentclass[sn-mathphys-num]{sn-jnl}

\usepackage{graphicx}%
\usepackage{multirow}%
\usepackage{amsmath,amssymb,amsfonts}%
\usepackage{amsthm}%
\usepackage{mathrsfs}%
\usepackage[title]{appendix}%
\usepackage{xcolor}%
\usepackage{textcomp}%
\usepackage{manyfoot}%
\usepackage{booktabs}%
\usepackage{algorithm}%
\usepackage{algorithmicx}%
\usepackage{algpseudocode}%
\usepackage{listings}%

\usepackage{multirow}
\usepackage{makecell}
\usepackage{color,soul}

\DeclareMathOperator*{\argmin}{arg\,min}

\theoremstyle{thmstyleone}%
\theoremstyle{thmstyletwo}%

\theoremstyle{thmstylethree}%

\begin{document}

\title[Article Title]{Robust 3D Point Clouds Classification based on Declarative Defenders}

\def\resp{\emph{resp. }}
\def\eg{\emph{e.g., }}
\def\ie{\emph{i.e., }}
\def\cf{\emph{c.f. }}
\def\etc{\emph{etc. }} 
\def\vs{\emph{vs. }}
\def\wrt{\emph{w.r.t. }}
\def\etal{\emph{et al. }}

\newcommand{\bfsection}[1]{\vspace*{0.1cm}\noindent\textbf{#1.}}








    
\author[1]{\fnm{Kaidong} \sur{Li}}

\author[1]{\fnm{Tianxiao} \sur{Zhang}}

\author[1]{\fnm{Cuncong} \sur{Zhong}}

\author[2]{\fnm{Ziming} \sur{Zhang}}

\author*[3]{\fnm{Guanghui} \sur{Wang}}\email{wangcs@torontomu.ca}

\affil[1] {\orgdiv{Department of Electrical Engineering \& Computer Science}, \orgname{University of Kansas}, \orgaddress{\street{1520 West 15th St.},
\city{Lawrence}, \postcode{66045}, \state{KS}, \country{USA}}}
\affil[2] {\orgdiv{Department of Electrical \& Computer Engineering}, \orgname{Worcester Polytechnic Institute}, \orgaddress{\street{100 Institute Rd.}, \city{Worcester}, \postcode{01609}, \state{MA}, \country{USA}}}
\affil[3] {\orgdiv{Department of Computer Science}, \orgname{Toronto Metropolitan University}, \orgaddress{\street{350 Victoria Street}, \city{Toronto}, \postcode{M5B 2K3}, \state{ON}, \country{Canada}}}


\abstract{
3D point cloud classification requires distinct models from 2D image classification due to the divergent characteristics of the respective input data. While 3D point clouds are unstructured and sparse, 2D images are structured and dense. Bridging the domain gap between these two data types is a non-trivial challenge to enable model interchangeability. Recent research using Lattice Point Classifier (LPC) highlights the feasibility of cross-domain applicability. However, the lattice projection operation in LPC generates 2D images with disconnected projected pixels. In this paper, we explore three distinct algorithms for mapping 3D point clouds into 2D images. Through extensive experiments, we thoroughly examine and analyze their performance and defense mechanisms. Leveraging current large foundation models, we scrutinize the feature disparities between regular 2D images and projected 2D images. The proposed approaches demonstrate superior accuracy and robustness against adversarial attacks. The generative model-based mapping algorithms yield regular 2D images, further minimizing the domain gap from regular 2D classification tasks. The source code is available at https://github.com/KaidongLi/pytorch-LatticePointClassifier.git.
}

\keywords{
Deep neural networks, 3D point cloud classification, mapping algorithm, adversarial defense.
}



\maketitle

\section{Introduction}\label{sec1}
Point clouds consist of data points sampled through light detection and ranging (LiDAR) sensors, playing a crucial role in various 3D vision tasks. LiDAR is widely employed in applications like autonomous driving because it provides accurate 3D measurements. With the advancements in 2D image tasks \cite{he2016deep, vaswani2017attention, li20202, li2021sgnet}, 3D point clouds also witnessed rapid development, primarily driven by the integration of deep neural networks (DNNs). However, significant work is required to design modules to adapt 3D data for DNNs. 

In recent years, several approaches have been proposed for 3D classification. \textit{Multi-view-based methods} \cite{su2015multi, mo2021stereo} projects 3D objects into multiple 2D data, which can either be 2D images or 2D features from different views. Although this approach can directly apply 2D models, it inevitably loses depth information. Moreover, the generated 2D images are usually scattered pixels, deviating from normal 2D images. \textit{Graph-based methods} \cite{wang2019dynamic, bruna2013spectral} employs graph representation to preserve spatial information like depth in 3D point cloud. Then graph convolution networks (GCNs) are utilized for end-to-end training. \textit{Voxel-based methods} \cite{wu20153d, zhou2018voxelnet} creates 3D voxels by dividing the space into occupancy grids. The features extracted in individual voxels are then gathered using 3D sparse convolutions. However, due to its computation complexity, its resolution is limited \cite{lu2020deep}. \textit{Point-based methods} \cite{qi2017pointnet, li2018pointcnn, xia2021building} extract point-wise features and aggregate them using customized modules. To effectively merge the traditionally separate branches of DNNs, it is essential to explore mapping algorithms that enable a seamless integration of 2D models.

\begin{figure}[t]
    \includegraphics[width=1\linewidth]{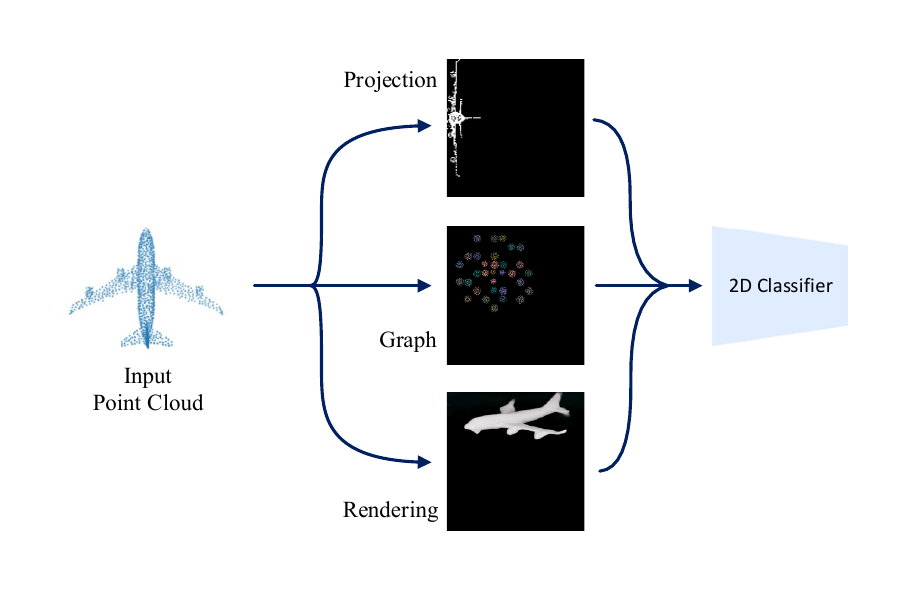}
    \caption{Illustration of mapping algorithms to transform 3D point clouds to 2D images.} 
    \label{fig:main}
\end{figure}

Recent work proposed by Li \etal \cite{li2022robust} showed that existing DNNs are vulnerable to adversarial attacks and proposed a new approach to 3D point cloud classification. Exploring correct attack algorithms \cite{xiang2019generating, ma2020efficient}, perfect attack results can be achieved on almost all of the state-of-the-art classifiers. Since most 3D classifier applications are safety-critical, like autonomous driving, it is imperative to adopt a robust classifier under adversarial attacks. 
In this paper, we extend the lattice point classifier (LPC) in \cite{li2022robust} and propose three additional mapping algorithms as shown in Figure \ref{fig:main}. We design a suite of experiments to evaluate and analyze their performance and defense abilities.

\textit{Direct projection.} Inspired by LPC \cite{li2022robust} which projects the 3D point cloud onto a lattice hyperplane, we introduce a simpler version to analyze the performance. Like LPC, both algorithms project 3D points onto a hyperplane and generate sparse pixels on the 2D plane. However, these geometric projection methods will lose depth information. \textit{Graph drawing.} Lyu \etal \cite{lyu2020learning} proposed a point cloud part segmentation model based on an efficient hierarchical graph-drawing algorithm to represent 3D point clouds using 2D grids. The spatial relationship can be preserved by graphs. We extend the algorithm to further improve its classifier performance and analyze its defense mechanisms.

These mapping algorithms are inspired by multi-view and graph methods. {The generated 2D images contain sparse projected pixels with the majority of the images filled with dark backgrounds, as shown in Fig.} \ref{fig:main}. To further align the generated 2D images with regular 2D images, we also introduce a generative model-based mapping algorithm to reduce the domain gap. 

The main contributions are summarized below.
\begin{itemize}
    \item We propose a family of structured declarative classifiers employing three distinct mapping algorithms. Through extensive experiments, we thoroughly analyze and enhance the performance of these classifiers. {The proposed projection classifier achieves state-of-the-art accuracy within the family of structured declarative classifiers. To the best of our knowledge, this is the first attempt to use graph-drawing and rendering techniques for 3D point cloud classification.}

    \item The proposed classifiers serve to narrow the gap between generated 2D images and real 2D images. We conduct experiments to showcase that our GAN-based classifier exhibits a minimal domain gap when compared to real 2D datasets.

    \item We analyze gradient propagation in declarative classifiers and investigate the defense mechanisms employed by these models. {Notably, we discovered a method to compromise the defense mechanism of the declarative defender. However, the basic projection and rendering classifiers demonstrate strong resilience to attacks.}

\end{itemize}

\section{Related Work}
\subsection{3D Point Cloud Classifications} 
Popular 3D point cloud classifications can be classified into four categories. \textit{Multi-view-based networks} \cite{su2015multi, yang2019learning, wang2023pst} project 3D point clouds onto one or multiple 2D hyperplanes to generate one or multiple 2D images to feed to a regular 2D neural networks. Then pooling algorithms can be applied to aggregate features. Voxel-based networks \cite{wu20153d, zhou2018voxelnet, le2018pointgrid} typically transform point clouds into volumetric occupancy grids, followed by the application of various classification techniques, such as 3D Convolutional Neural Networks (CNNs). Graph-based networks \cite{wang2019dynamic, lyu2020learning, fu2021robust} often represent point clouds as graphs. {Unlike multi-view methods, where depth information is inevitably lost during projection, graph-based methods preserve spatial information by utilizing structures such as K-nearest neighbor (KNN) or adjacency graphs.} These graph representations are then leveraged for training Graph Convolutional Networks (GCNs). On the other hand, \textit{Point-based networks} \cite{qi2017pointnet, qi2017pointnet++, li2018pointcnn} directly take individual points as input, applying multi-layer perceptrons (MLPs) to extract point-wise features, which are subsequently aggregated to generate a comprehensive representation for the entire point clouds. Then PointNet++ \cite{qi2017pointnet++} applies farthest point sampling (FPS) to group points at different scales to gather hierarchical features.

\subsection{3D Point Cloud Attacks and Defenses} 
Deep learning-based models are proven to be vulnerable to attacks \cite{szegedy2013intriguing,  kurakin2018adversarial, xiang2019generating}. Adversarial attacks aim to trick the DNN models by slightly modifying the input data, in a way that is indiscernible by human eyes. Many gradient-based methods can be applied to 3D point clouds. We categorize attacks based on how the input point clouds are modified. \textit{Point perturbation} \cite{xiang2019generating, liu2019extending, ma2020efficient, kim2021minimal} is the most common type. It attacks the point clouds by moving all (or part of) the points by a small amount. The attack efficiency can be demonstrated by the amount of perturbation, which is often measured by perturbation $L_2$ norm. Advanced versions \cite{xiang2019generating, liu2019extending} will constrain the number of perturbations in the optimization process. \textit{Point addition} \cite{xiang2019generating, wicker2019robustness} tries to trick models by adding extra points. These added points can be independent or sampled from targeted class objects. Then, slight perturbations, scaling and rotations are added to the extra points to increase attack efficiency. In contrast, \textit{point dropping} \cite{zheng2019pointcloud, wicker2019robustness, } is a technique where adversarial attackers select critical points from each input point cloud and remove them to deceive the classifier. 

Nevertheless, this operation lacks differentiability. A significant portion of research endeavors is dedicated to transforming point-dropping into an alternative action that facilitates gradient propagation. For example, Zheng \etal \cite{zheng2019pointcloud} proposed to view it as relocating points to the cloud center. Other categories of attacks also have seen great success. LG-GAN \cite{zhou2020lg} is based on generative adversarial networks (GANs). Backdoor attacks \cite{xiang2021backdoor, zhang2022towards} poison the model training process by inserting a backdoor pattern so that the victim models will be tricked if the backdoor pattern is presented during inference. \cite{zhao2020isometry, dong2022isometric} proposed attacks that force constraints on isometric property. 

Adversarial defense in 3D point clouds is important because many applications of 3D models are safety-critical. The defense approaches can be summarized into the following categories. (1) Statistical Outlier Removal (SOR) \cite{rusu2008towards} tries to defend by removing points to make a smooth surface. In addition, SOR is non-differentiable, obfuscating gradients for defenders. DUP-Net \cite{zhou2019dup} and IF-Defense \cite{wu2020if} both employ SOR and append a geometric-aware network to refine the point clouds. (2) Random sampling exhibits robustness against adversarial attacks \cite{yang2019adversarial}. PointGuard \cite{liu2021pointguard} introduced a classification method by randomly sub-sampling point clouds multiple times, and making predictions based on the majority votes. (3) Data augmentation emerges as another effective defense mechanism. Zhang \etal \cite{zhang2021art} demonstrated robustness by simply doing random permutation. Point-CutMix \cite{zhang2021pointcutmix} swaps some points between the training pairs to generate new data. (4) Another approach is using structured robust declarative classifiers \cite{li2022robust}. Some other defenders employ the diffusion model to perform point purification. Ada3Diff \cite{zhang2023ada3diff} estimates distortion based on points distance to their best-fitting plane. Li \etal \cite{li2022improving} proposed a method using a graph network together with a data augmentation method to ensure robustness. 

\subsection{Graph Drawing} 
The goal of graph drawing is to represent original 3D point clouds using graph $\mathcal{G=(V, E)}$ and draw the graph on a low-dimensional space \cite{schnyder1990embedding, didimo2019survey, lyu2020learning}. By different graph properties (e.g., spatial relationships among vertices and edges), graph representations can be divided into many families. For example, $k$-planar drawing \cite{pach1997graphs} limits the number of edge crossings, and RAC drawings \cite{didimo2011drawing} will only have perpendicular crossings on their edges. Lyu \etal \cite{lyu2020learning} proposed a Delaunary triangulation-based \cite{delaunay1934sphere} graph drawing method for 3D segmentation. To directly apply normal 2D networks, a novel hierarchical approximation algorithm is designed to generate normal images from graph representation while preserving local information in point clouds.

\begin{figure}[t]
    \includegraphics[width=1\linewidth]{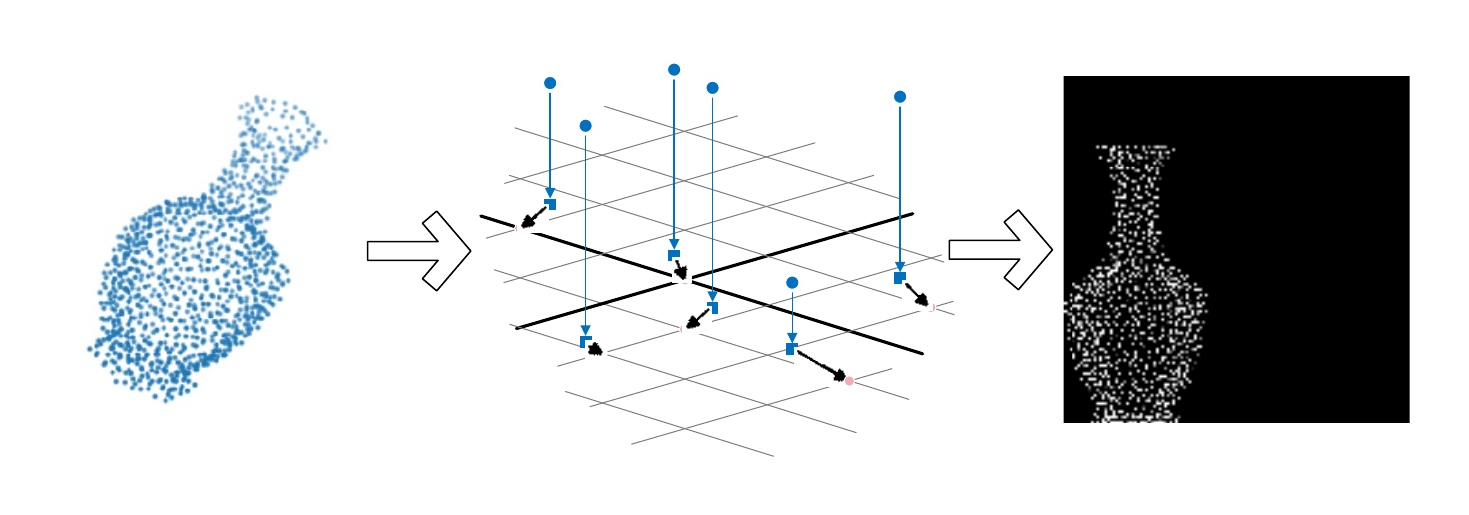}
    \caption{Projection Classifier. Input point clouds are projected onto the $x-y$ axis plane by removing $z$-axis information. The coordinates of projected points are rounded using the floor function. } 
    \label{fig:proj}
\end{figure}

\subsection{Reconstruction and Rendering}
Reconstruction surface from point clouds is an important and long-standing research direction in computer graphics, bridging the gap between points and images. A common approach is to form triangulations to compute a volume tetrahedralization \cite{bernardini1999ball, cazals2004delaunay}. But these methods tend to create undesirable holes especially when points get noisy \cite{metzer2022z2p}. Implicit surface reconstruction \cite{nagai2009smoothing, kazhdan2013screened} can handle noise better at the cost of requiring more computational resources. Katz \etal \cite{katz2007direct} proposed a method to reconstruct the surface only from one single view using hidden point removal.
The rapid development of DNNs shines light on the potential of bypassing manually designed priors. Hanocka \etal \cite{hanocka2020point2mesh} proposed a DNN-based method to deform an initial mesh to shrink-wrap around an input point cloud. A differentiable surface splatting algorithm is proposed by Yifan \etal \cite{yifan2019differentiable} to update point locations and normals. Z2P \cite{metzer2022z2p} proposed a method to view the rendering as a point-depth-map-to-image problem, which employs a modified U-Net \cite{ronneberger2015u} to render 2D images, achieving robust performance under noises and non-uniformly sampling.

\begin{figure}[t]
    \includegraphics[width=1\linewidth]{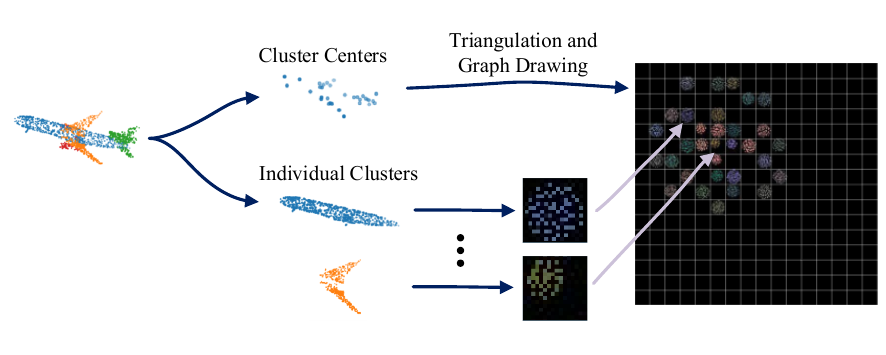}
    \caption{Graph Drawing Classifier. Input point clouds are divided into 32 clusters using balanced KMeans clustering \cite{lyu2020learning}. Delaunay triangulation \cite{delaunay1934sphere} is applied on the 32 cluster centers and within each individual cluster, generating 2-level graphs. Then the top-level graph is mapped to a $16 \times 16 $ grid, where each cluster center occupies a grid cell. Then each grid cell is filled with a lower-level $16 \times 16$ grid, obtained from within-cluster graphs, yielding the final $256 \times 256$ image.} 
    \label{fig:graph}
\end{figure}

\subsection{Multi-modal Large Language Models} 
Multi-modal Large Language Models (MLLMs) are designed to achieve a broad understanding across various modalities, encompassing audio \cite{huang2023audiogpt}, image \cite{radford2021learning}, point cloud \cite{xue2023ulip}, and more. The underlying idea is that incorporating language modality can enhance a model's ability to comprehend high-level interactions within diverse input data. Within this domain, one category of MLLMs employs language as a means to interact with other modalities \cite{li2019visualbert, xu2023pointllm}. These models showcase versatile capabilities, particularly when provided with text inputs as prompts. Another category of MLLMs adopts individual encoders for each modality, aligning features from diverse modalities \cite{radford2021learning, chen2024superlora,}. These models demonstrate notable zero-shot generalization capabilities. Noteworthy examples include CLIP \cite{radford2021learning}, which has inspired a myriad of novel applications. For instance, PointCLIP \cite{zhang2022pointclip} leverages CLIP for zero-shot 3D multi-view classification, while Xue \etal \cite{xue2023ulip} utilize CLIP's multi-modal features to supervise a 3D point cloud encoder, aligning three modalities. Given CLIP's proven robustness across multiple tasks, we leverage CLIP to assess the domain gap of generated 2D images when compared to typical 2D images.

\begin{figure*}[t]
    \centering
    \includegraphics[width=1.0\linewidth]{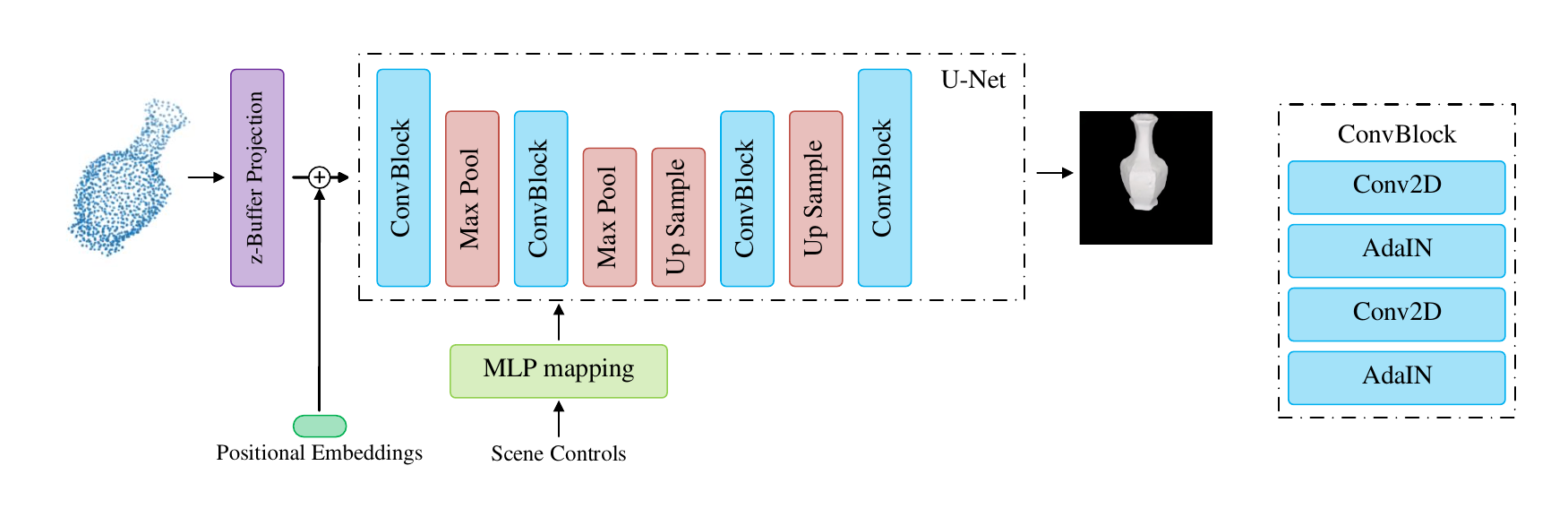}
    \caption{Rendering Classifier. Input point clouds are converted to a 2D depth map using z-buffer projection. Positional embeddings are appended to the depth map to boost global context \cite{metzer2022z2p}. Then the generated features are fed to a modified U-Net \cite{ronneberger2015u} to produce final rendered images. In the modified U-Net, the normalization layer is replaced with a style-based adaptive instance normalization (AdaIN) \cite{karras2019style}, which also takes scene control features to influence output image style. } 
    \label{fig:render}
\end{figure*}

\section{Mapping Algorithms}
The section describes the three mapping algorithms that can be used to map 3D point clouds into 2D space.

\subsection{Basic Projection Classifier}
The Lattice Point Classifier (LPC) was introduced as an implementation of a robust declarative classifier (RDC) with robust defense capabilities against gradient-based attacks \cite{li2022robust}. However, LPC involves a sophisticated mapping algorithm. It first projects 3D point clouds onto a lattice hyperplane \cite{adams2010fast}, which is populated with a triangular lattice. Subsequently, each projected point is splatted onto the three enclosing triangular lattice vertices. In this context, we delve into a more straightforward mapping algorithm for 3D-to-2D projection by simply dropping one dimension of the data as shown in Figure \ref{fig:proj}. This basic projection serves as a baseline for performance comparison, and the defensive behavior of RDC is scrutinized in conjunction with this basic projection.

\subsection{Graph Drawing Classifier}
Both basic projection and permutohedral lattice projection unavoidably lead to the loss of depth information perpendicular to the 2D hyperplanes. In contrast, graph representations showcase the capability to preserve 3D spatial information, rendering them an optimal choice for the mapping algorithm. 

Given the graph $\mathcal{G=(V, E)} = h(\mathbf{X})$, where $\mathbf{X} = \{ x_i \in \mathbb{R}^3 \}$ denotes input point cloud, the function $h: \mathbf{X} \rightarrow \mathcal{G}$ transform input point clouds to graph representations. Graph drawing is originally employed to transform a graph $\mathcal{G}$ into continuous drawings. To facilitate the utilization of standard 2D neural networks, we want to find a function $f: \mathcal{G} \rightarrow \mathbb{Z}^2$ that converts graphs to 2D integer coordinates. Inspired by the graph drawing segmentation model \cite{lyu2020learning}, we adopt an efficient graph drawing classifier. This method reduces computational costs by creating two hierarchical levels of graphs and incorporating balanced KMeans clustering. The outlined approach is illustrated in Fig. \ref{fig:graph}.

\bfsection{Two-level graph drawing}
The computational complexity of Delaunay triangulation is $\mathcal{O}(n^{ \lceil d/2 \rceil})$ \cite{amenta2007complexity}, where $n$ is the number of data points and $d$ is the dimension of each data point. For 3D point cloud, the complexity becomes $\mathcal{O}(n^2)$.
Thus, instead of directly applying graph drawing on all the 3D points $\mathbf{X}$, we employ a two-level method that divides $\mathbf{X}$ into 32 clusters and creates a two-level hierarchical data, $\mathbf{C}=\{ \mu_1,\mu_1, \cdots, \mu_{32}  \}$ and $\mathbf{S}=\{ S_i\}$, where $ S_i=\{ x_j \in \mathbb{R}^3\}$ and $\mu_i$ is the center of cluster $S_i$. The Delaunay triangulation \cite{delaunay1934sphere} is then applied to the point set $\mathbf{C}$ and $S_i$. This approach will reduce computational complexity.

\subsection{Rendering Classifier}
As shown in Fig. \ref{fig:main}, the images generated from the above mapping algorithm still have distinctive visual artifacts. The rendering-based method can generate realistic 2D images by reconstructing object surfaces, which can further reduce the domain gap from regular 2D tasks. Z2P \cite{metzer2022z2p} is a lightweight DNN-based rendering algorithm. The pipeline of the mapping is shown in Fig. \ref{fig:render}. It first generates a z-buffer projection from the original point cloud, and then renders the 2D image through a modified U-Net. 

\bfsection{z-buffer Projection} Like the basic projection method, it projects points onto a 2D hyperplane, obtaining a pixel-point pair $(x_k, I(i, j))$, where $x_k$ is a 3D point and $I(i, j)$ is the corresponding pixel value at image coordinate $(i, j)$. The pixel value is determined by the distance of the 3D point from the image plane $d$. The pixel value can be calculated by $I(i, j) = e^{-(d-\alpha)/\beta}$, meaning the farther the points are, the darker they appear on the depth image. The pixel value is shared within a $3\times 3$ pixel window centered at $I(i, j)$. The pixels without a corresponding point are given a value $0$. 

\bfsection{Adaptive Instance Normalization (AdaIN)} To enhance control over the colors and shadows of the rendered images, we replace the original normalization layers with AdaIN \cite{karras2019style}. In addition to the normal DNN features $\mathbf{f}$ of size $H \times W \times C$, AdaIN takes a scene-control vector $w$ as input. Each AdaIN layer contains two affine transformations, denoted as $A_s$ and $A_b$, which are utilized to compute styles $y_s = A_s \cdot w$ and $y_b = A_b \cdot w$. The output of AdaIN is determined by
\begin{align}\label{eqn:ada}
    \text{AdaIN}(f_i, w) = y_{s,i} \frac{f_i - \mu(f_i)}{\sigma(f_i)} + y_{b, i}
\end{align}
where $\mu(f_i)$ and $\sigma(f_i)$ are the mean and standard deviation (std) at each feature channel $f_i$.

\section{Experiments}
\subsection{Dataset}
\textbf{ModelNet40} \cite{wu20153d}: The dataset comprises clean synthetic 3D objects generated from CAD models, encompassing 40 distinct categories,  with a total of 12,311 objects. These objects are split into 9,843 training samples and 2,468 testing samples. In our experimental setup, 3D points are uniformly sampled from the mesh surface, adhering to a specified methodology in \cite{qi2017pointnet, xiang2019generating}. 

\subsection{Training Settings}
We perform classification training on the ModelNet40 dataset. For all training tasks, we run 200 training epochs using Adam optimizer \cite{kingma2014adam} with 0.0001 decay rate and $(0.9, 0.999)$ beta values. A simple step learning rate scheduler is used with a step size of 20 and a starting learning rate of 0.001 unless otherwise specified. The point clouds are evenly sampled from the object surface following the practice in \cite{qi2017pointnet}. During training, the points are augmented by random point dropout, scale, and shift. The max dropout rate is 0.875. The random scale rate is set from 0.8 to 1.0. The max random shift distance is 0.1. When random rotation is enabled, the max random rotation angle is $\pi$. No augmentation techniques are applied during the inference.

\bfsection{Basic Projection Classifier} We scale the point clouds to generate $456 \times 456$ images. During data augmentation, points projected outside the image will be set to pixel $I(0, 0)$. 

\bfsection{Graph Drawing Classifier} Using the balanced KMeans \cite{lyu2020learning}, the original input point clouds are divided into $K=32$ clusters. The balanced KMeans clustering algorithm initially generates unbalanced clusters $h$, and then starts to reduce the number of points in oversize clusters. A cluster is considered oversized when the number of points $|h|> \alpha \cdot | \mathbf{X} | / K$ where $| \mathbf{X} |$ is the total number of points in input point clouds, and $\alpha$ is set to 1.2. For the graph drawing, both the lower-level and top-level grid sizes are $16 \times 16$, making the final graph drawing output to be $256 \times 256$.

\bfsection{Rendering Classifier} The classifier undergoes a two-step training process. In the first step, the rendering model is trained for 10 epochs with a learning rate of $3 \times 10^{-4}$. Throughout the z-buffer projection, the splat window size is set to 3. During inference for scene control, the generated object color RGB values are fixed at $(255, 255, 255)$, and lighting is positioned at the origin point for simplicity. The resolution of the generated rendered 2D image is set to $313 \times 313$. The rendering training data is generated following \cite{metzer2022z2p} using \cite{blendersite}.

\begin{table}
\centering
\setlength{\tabcolsep}{12pt}
\renewcommand{\arraystretch}{1.4}
\caption{Zero-shot classification accuracy (\%) using CLIP \cite{radford2021learning}. Text inputs are a list of the ModelNet40 \cite{wu20153d} class names. Image inputs are 2D images generated by three tested projection algorithms. }
\begin{tabular}{  c | c | c }
    \toprule
                            & Instance Accuracy & Class Accuracy \\
    \midrule
    LPC \cite{li2022robust} &     2.64          &    3.75        \\
    Basic Projection       &     5.17          &    5.43        \\
    Graph Drawing           &     2.32          &    1.95        \\
    Rendering               & \textbf{27.08}    & \textbf{23.78} \\
    \bottomrule

\end{tabular}
\label{table:clip_accuracy}
\end{table}

\begin{table}
\centering
\setlength{\tabcolsep}{12pt}
\renewcommand{\arraystretch}{1.4}
\caption{Classification accuracy (\%) using test projection algorithms. }
\begin{tabular}{  c | c | c }
    \toprule
                            & Instance Accuracy & Class Accuracy \\
    \midrule
    LPC \cite{li2022robust} &            89.51   &         86.30  \\
    Basic Projection       &    \textbf{91.02}  & \textbf{88.42} \\
    Graph Drawing           &            84.97   &   81.21    \\
    Rendering               &            88.30   &   85.78    \\
    \bottomrule

\end{tabular}
\label{table:accuracy}
\end{table}

\begin{table}
\centering
\renewcommand{\arraystretch}{1.4}
\caption{Ablation study for rendering classification. The table shows the inference accuracy ($\%$) with different setups. The mean and std of the rendered normalization are calculated from the training set of rendered 2D images. When upright is not checked, the model is trained with default point cloud orientation. }
\begin{tabular}{  c | c  c  c  c | c  c }
    \toprule
                        
    Backbone                        & \thead{Learnable\\Backbone} & \thead{Rendered\\Normalization} & Upright & \thead{Random\\Rotation} & \thead{Instance\\Accuracy} & \thead{Class\\Accuracy} \\
    \hline
    \hline
    ViT-1 &            & & & & 78.53 & 74.64  \\
    ViT-2 & \checkmark & & & & 76.53 & 71.53  \\
    ViT-3 & & \checkmark & & & 77.46 & 73.53  \\
    \midrule
    ResNet50-1      &            &            & & & 67.50 & 60.15      \\
    ResNet50-2      &            & \checkmark & & & 67.06 & 59.59  \\
    ResNet50-3      & \checkmark & \checkmark & & & 56.81 & 49.86  \\
    \midrule
    EfficientNet-B5 & \checkmark & & & & 83.65 & 79.31       \\
    EfficientNet-B5 & \checkmark & \checkmark & & & 84.29 & 80.54 \\
    EfficientNet-B5  & \checkmark & & \checkmark & & 86.61 & 82.86 \\
    EfficientNet-B5 & \checkmark & \checkmark & \checkmark & & 86.90 & 83.70 \\
    EfficientNet-B5 & \checkmark & \checkmark & \checkmark & \checkmark & \textbf{88.30} & \textbf{85.78} \\
    \bottomrule

\end{tabular}
\label{table:gan_ablation}
\end{table}

\subsection{Domain Gap}
A crucial factor for the application of regular 2D classifiers is minimizing domain gaps between generated and regular images. To quantify this domain gap, we leverage well-established foundation models \cite{radford2021learning, jia2021scaling} that have been trained on extensive general domain data with over 15 million images. These models have demonstrated notable zero-shot classification accuracy across diverse datasets. For instance, CLIP \cite{radford2021learning} achieves a $76.2\%$ zero-shot classification accuracy on the extensive ImageNet dataset \cite{deng2009imagenet}. The inherent alignment of text-image features in these pre-trained models simplifies their adaptation to diverse classification tasks across datasets with varying categories.

We use CLIP \cite{radford2021learning} to run zero-shot classification on the generated 2D image from LPC \cite{li2022robust} and our three declarative classifiers. The image backbone is ViT-B/32 \cite{dosovitskiy2020image} which takes images of resolution 224$\times$224. We directly utilize CLIP's image normalization mean and std, $(0.4814, 0.4578, 0.4082)$ and $(0.2686, 0.2613, 0.2758)$, for fair comparison. The text inputs are the category names of each ModelNet40 class, [```airplane", ``bathtub", ``bed", ..., ``wardrobe", ``xbox"].

\subsubsection{CLIP Zero-shot Accuracy} \label{sect:zero_shot_accu}
The zero-shot classification results are shown in Table \ref{table:clip_accuracy}. We have three observations from the results. (i) Graph drawing 2D images have the lowest accuracy, hovering around the same level of total random prediction accuracy $\frac{1}{NUM\_CLASS} = 1/40 = 2.5\%$. It is not surprising when we take a look at the visualization of graph drawing images depicted in Fig. \ref{fig:vis_main}. However, considering that the primary goal of graph representation is to encode point clouds into latent features rather than human-understandable images, this approach remains promising, as evidenced by the $85.58\%$ classification accuracy. (ii) The two projection-based methods (LPC and basic projection classifier) still demonstrate a substantial domain gap. The sparse enabled pixels in Fig. \ref{fig:vis_main} are different from regular dense images. However, the accuracy difference between LPC (2.64\%) and basic projection (5.17\%) indicates that point of view can significantly affect classifier performance. This point is further validated in subsequent experiments on rendering classification in the ablation study. (iii) Despite rendering classifier 2D images achieving a noteworthy reduction in domain gap compared to other mapping algorithms, other classifiers do not achieve satisfactory CLIP zero-shot performance as shown in  Table \ref{table:clip_accuracy}.

\begin{figure}[t]
    \centering
    \includegraphics[width=1\linewidth]{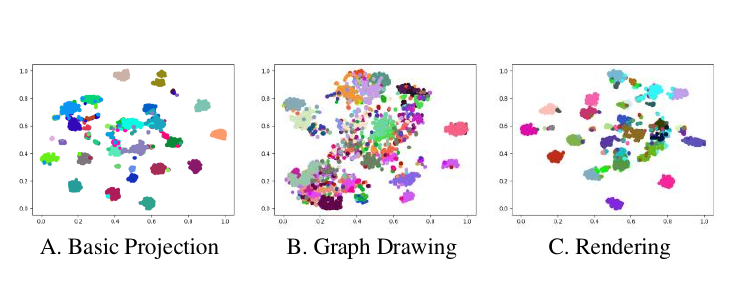}
    \caption{Visualization of 2,468 ModelNet40 \cite{wu20153d} test point clouds using t-SNE \cite{van2008visualizing}. Each plot is based on the DNN features generated by corresponding mapping algorithms before linear layers. } 
    \label{fig:tsne}
\end{figure}

\subsection{Classification Accuracy}

\begin{figure*}[t]
    \centering
    \includegraphics[width=1\linewidth]{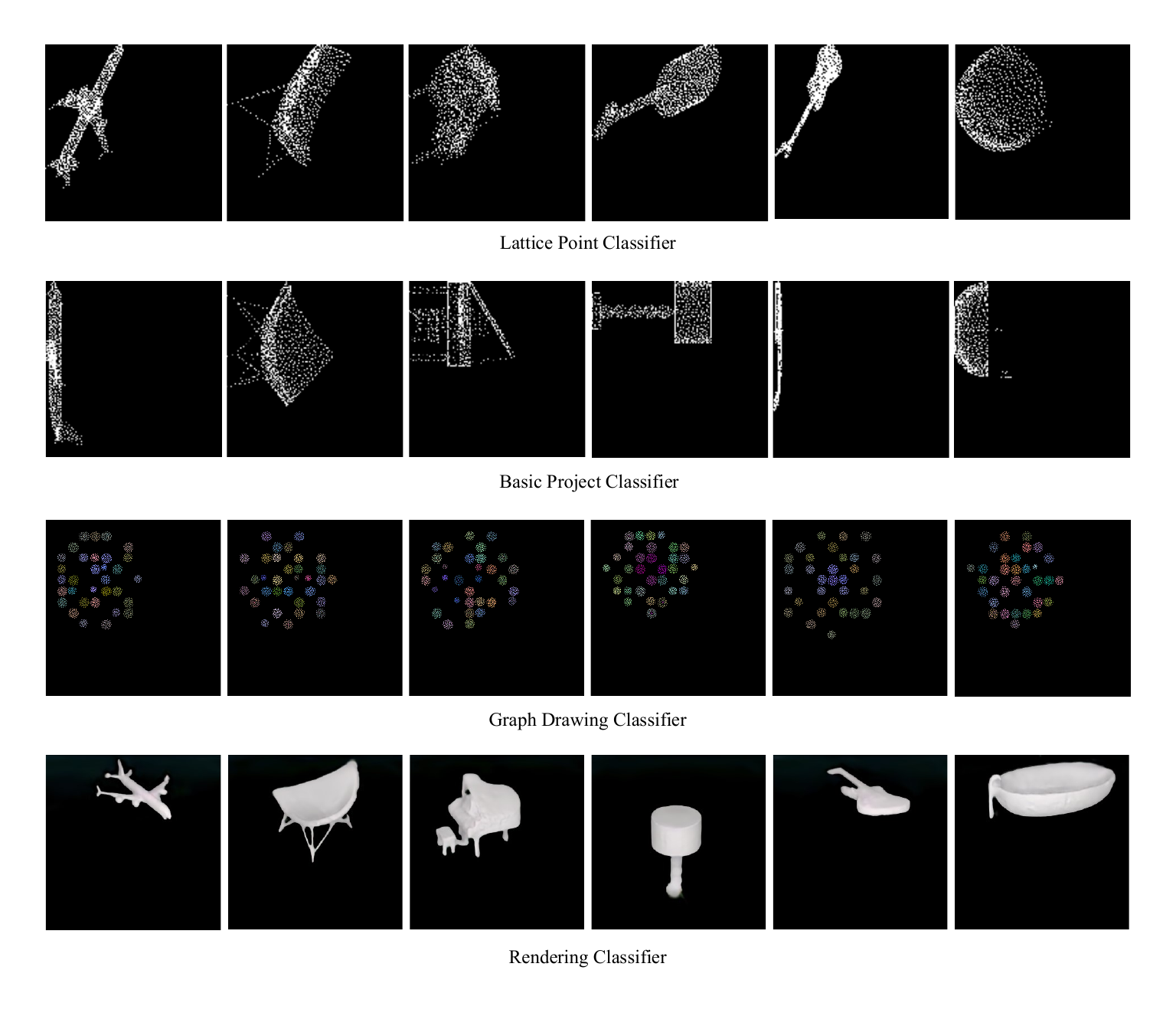}
    \vspace{-7mm}
    \caption{Visualization of 2D images generated by LPC, basic projection, graph drawing, and rendering classifiers.} 
    \label{fig:vis_main}
\end{figure*}

For all four classification models, we employ EfficientNet-B5 \cite{tan2019efficientnet} as the 2D backbone networks, and the results are summarized in Table \ref{table:accuracy}. The basic projection model achieves the highest test accuracy at $91.02\%$, outperforming LPC \cite{li2022robust}. As visualized in Fig. \ref{fig:vis_main}, LPC and basic projection essentially represent high-to-low-dimensional projections from different viewpoints, which significantly influences the classification performance. This observation is further validated in subsequent ablation studies. The performance of graph drawing demonstrates that DNNs can be trained to learn distinct data representations without any modifications. Despite generating more realistic 2D images, the surprising outcome is that rendering falls behind the two projection methods. We conduct a more in-depth analysis of the model's performance through t-SNE visualization \cite{van2008visualizing} in subsequent sections.

t-SNE \cite{van2008visualizing} is a technique to visualize high-dimensional data to two or three-dimensional data, offering insights into data separation. In our analysis, we input feature vectors from each mapping algorithm, obtained before linear layers, into the t-SNE algorithm. The resulting visualization is depicted in Fig. \ref{fig:tsne}. It is evident that graph drawing's clusters are more ambiguous, with less separation and blurred boundaries. Comparatively, upon closer inspection, the t-SNE representation for basic projection reveals more defined clusters. Each class forms a smaller, more separated cluster compared to the graph generated by rendering. Moreover, in the central region of the t-SNE visualizations, rendering exhibits more ambiguous data points, providing an explanation for its lower accuracy.

\begin{figure}[t]
    \centering
    \includegraphics[width=1\linewidth]{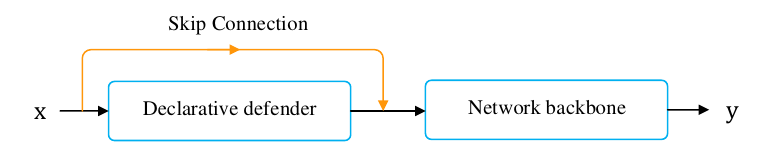}
    \caption{Gradient skip connection. The normal declarative classifier is the module without the orange skip connection. Declarative defenders will block back-propagation using implicit gradients. In graph drawing, 3D point coordinates are directly assigned as image intensities, creating a skip connection and bypassing declarative defender.} 
    \label{fig:defense_skip}
\end{figure}

\begin{table}[h]
\centering
\renewcommand{\arraystretch}{1.4}
\caption{Instance accuracy (\%) of basic projection classifier with different settings. }
\begin{tabular}{  c c c | c  }
    \toprule
    backbone & \# linear layers & 2D dimension                        & Instance Accuracy  \\
    \midrule
    ViT-L         &  1 & 224 & 55.11     \\
    ViT-L         &  3 & 224 & 79.72     \\
    ViT-L         &  3 & 448 & 78.37      \\
    ResNet        &  3 & 448 & 77.88      \\
    \bottomrule

\end{tabular}
\label{table:bpc_ablation}
\end{table}

\subsection{Ablation Study}
\bfsection{Basic Projection Classifier} {We ran experiments with different settings, as shown in Table {\ref{table:bpc_ablation}}. Comparing the first two experiments, we observe that additional linear layers improve classification accuracy. Another observation, from comparing the second and third rows, is that higher 2D resolution does not necessarily lead to better performance. Increasing the number of pixels in the converted 2D images results in more sparsity, which slightly reduces accuracy for the ViT-L backbone.}

\bfsection{Rendering Classifier} We evaluate the contribution of different setups on our rendering classifier. Many applications have proven the power of foundation models trained on large amounts of data \cite{radford2021learning, wang2022clip, ramesh2021zero, saharia2022photorealistic}. We employ two different CLIP \cite{radford2021learning} image encoders (ViT \cite{dosovitskiy2020image} and ResNet50 \cite{he2016deep}), which produce 768 channel and 1024 channel image features. We append 3 fully connected layers after the foundation image encoders, with 4096 as the fully connected layers' hidden channel dimension. The ablation results can be found in Table \ref{table:gan_ablation}.

\subsubsection{Learnable Backbone Weights}
For the foundation model encoders, we run training with the encoder weights frozen or trainable to study the effects. From the foundation ViT experiments (1 and 2) and ResNet50 experiments (2 and 3), we can see making the foundation encoder learnable will reduce the accuracy. Specifically, the accuracy decreases by $2\%$ to $76.53\%$ for ViT and $10\%$ to $56.81\%$ for ResNet50. Examining the training data accuracy, we observe that the frozen and learnable backbones using ResNet50 have $58.93\%$ and $56.38\%$ accuracy respectively. The small training accuracy gap indicates learnable foundation encoder tends to overfit the data. 

{Foundation models are typically 10 times larger than regular models. While the ModelNet40 dataset {\cite{wu20153d}} contains only around 9,000 samples, in contrast, a standard image classification dataset like ImageNet {\cite{deng2009imagenet}} has over one million samples. Moreover, the foundation encoder in CLIP is trained on 400 million text-image pairs {\cite{radford2021learning}}. Due to the limited number of training samples, fine-tuning on smaller datasets often results in sub-optimal performance, leading to issues such as overfitting and catastrophic forgetting.}

\subsubsection{Normalization} We conducted a comparison between different normalizations: CIFAR-100 normalization and rendered image normalization with mean values of (0.1136, 0.1135, 0.1128) and standard deviations of (0.2729, 0.2719, 0.2761) during training. Utilizing EfficientNet-B5 \cite{tan2019efficientnet}, we observed an improvement in accuracy by $0.64\%$ and $0.29\%$. However, this conclusion does not extend to foundation encoders. Specifically, with the foundation ViT backbone, the calculated normalization results in a decrease in accuracy by $1.07\%$. We suggest that foundation models might have a sufficient volume of data, allowing them to adapt to various data distributions.

\subsubsection{Observation View Point} As discussed in Section \ref{sect:zero_shot_accu}, the viewpoint might have a significant impact on classification accuracy. As shown in Table \ref{table:gan_ablation}, rotating object upright significantly improves classification accuracy. When CIFAR-100 normalization is used, the inference accuracy is boosted by $2.96\%$, and with accurate normalization, the improvement is $2.61\%$. 

\subsubsection{Random Rotation} The rendering classifier training is carried out in three steps: initially to train the rendering model, then to generate rendered images, and finally training on 2D images to save time. Consequently, during the training of the 2D image classifier, it only has exposure to the 3D object from a single viewpoint. To enhance the dataset, four additional sets of rendered images are created through the rotation of the original input point clouds. Specifically, we rotate $\pm \frac{\pi}{9}$ along $x$-axis and $\pm \frac{\pi}{9}$ along $z$-axis separately to generate the four sets of rendered images. This data augmentation strategy contributes to an increase in inference accuracy by $1.4\%$.

\subsection{Learning Rate}
{We also conducted experiments to show the sensitivity of our models to different learning rates, as shown in the following Table {\ref{table:lr}}. }

\begin{table}[h]
\centering
\renewcommand{\arraystretch}{1.4}
\caption{Instance accuracy (\%) with different starting learning rate on rendering classifier using ResNet50 backbone. }
\begin{tabular}{  c | c  }
    \toprule
    lr                        & Instance Accuracy  \\
    \midrule
    0.01        &            25.24     \\
    0.001       &    67.50            \\
    0.0001      &    65.81      \\
    \bottomrule

\end{tabular}
\label{table:lr}
\end{table}

{In all experiments, we used the Adam optimizer with a step learning rate scheduler and a step size of 20. For the rendering classifier, the model quickly diverged during training with a learning rate of 0.01, while a learning rate of 0.0001 was too small, causing the model to get stuck in local minima.}



\subsection{Defense Experiment Settings}
To test the defense performance of each mapping algorithm, we run FGSM \cite{goodfellow2014explaining} similar to LPC \cite{li2022robust} under various gradient-based attacks. The FGSM attack is applied on the entire test set from ModelNet40 \cite{wu20153d}. The attack learning rate of FGSM is set to 0.1.

\begin{table}
\centering
\setlength{\tabcolsep}{4pt}
\renewcommand{\arraystretch}{1.4}
\caption{Defense Performance (\%). }
\begin{tabular}{  c | c c | c }
    \toprule
        & \multicolumn{2}{c|}{Accuracy} & Attack Success Rate \\
        & No Attack & FGSM \cite{goodfellow2014explaining} & \\
    \midrule
    PointNet \cite{qi2017pointnet} & 90.15 & 45.99 & 48.99 \\
    LPC \cite{li2022robust} & 89.51 & 89.51 & \textbf{0.00} \\
    \midrule
    Basic Projection & \textbf{91.02} & \textbf{91.02} & \textbf{0.00} \\
    Graph Drawing&  84.97& 24.76& 70.86 \\
    Rendering& 88.30& 88.30& \textbf{0.00}\\
    \bottomrule

\end{tabular}
\label{table:defense_accu}
\end{table}

\subsection{Defense Performance}
The attack results are listed in Table \ref{table:defense_accu}. From the results, we can observe: (i) Basic projection and rendering classifier exhibit similar defense behavior as LPC \cite{li2022robust}. They are all robust under gradient-based attackers. (ii) Not all robust declarative classifiers are guaranteed to be robust under attackers. Graph drawing classifier is still vulnerable to adversarial attacks. 

We designed more experiments to further investigate the factors that influence defense performance. An illustration of the original declarative classifier can be found in Fig. \ref{fig:defense_skip} without the orange connection. The classifier can be formulated as
\begin{equation}\label{eqn:declarative}
    y = g(h(\mathbf{x}); \omega) 
\end{equation}
where $h(\mathbf{x}) = \argmin_{z\in \mathcal{Z}} f(x,z;\theta)$, $g(\Tilde{\mathbf{x}}; \omega)$ is the network backbone, which takes the generated images $\Tilde{\mathbf{x}} = h(\mathbf{x})$ as input, and $h(\mathbf{x})$ is the declar ative defender. In a gradient-based attack, the attack gradient can be easily back-propagated through $g(\Tilde{\mathbf{x}}; \omega)$. The declarative defender $h(\mathbf{x})$ will block gradient propagation using implicit gradients. 

In the context of graph drawing, a particular operation functions as a skip connection, compromising the defense capability. Each point in 3D space is mapped to a corresponding pixel in the resulting 2D image, and the coordinates of each point are then assigned as pixel intensities. This establishment of a skip connection circumvents the declarative defense mechanism, as illustrated in Fig. \ref{fig:defense_skip}. To validate this hypothesis, we conducted experiments where we leaked 3D coordinates as pixel intensity in the generated 2D images for the basic projection classifier. The attack success rate increased to $35.7\%$, effectively defeating the defense capability.

\section{Conclusion}
In this paper, we have investigated three distinct mapping algorithms as declarative defenders for 3D point cloud classification, demonstrating the robust performance of declarative nodes through diverse 3D-to-2D mapping techniques. {The majority of the proposed declarative defenders exhibit resilience against adversarial attacks, with the basic projection classifier achieves the best performance. Through the use of a rendering classifier, we discover that the observation viewpoint and random rotation can substantially impact classification accuracy.} Notably, upon observing the diminishing defense capability in the graph drawing classifier, we conclude that the presence of a skip connection circumvents implicit gradients, thereby defeating the defense capability.

There is room for improvement in the rendering classifier. Currently, the scene control parameters are manually set, mainly due to the scarcity of training data with actual object textures. A potential avenue for future work could involve designing a module to learn texture priors from the point cloud. This approach aims to generate more realistic 2D images by incorporating learned texture information.

\section*{Acknowledgement}
This research was partly funded by the Natural Sciences and Engineering Research Council of Canada (NSERC) and the Rogers Cybersecure Catalyst Fellowship Program.

\section*{Statements and Declarations}


\bmhead{Conflicts of interest} The authors have no conflicts of interest to declare that are relevant to the content of this article.

\bmhead{Data availability and source code} The study was evaluated using public datasets as cited in the paper. The source code is available at https://github.com/KaidongLi/pytorch-LatticePointClassifier.git.

\bibliography{cite} 

\end{document}